\documentclass[lettersize,journal]{IEEEtran}
\usepackage{amsmath,amsfonts}
\usepackage{algorithmic}
\usepackage{algorithm}
\usepackage{array}
\usepackage[caption=false,font=normalsize,labelfont=sf,textfont=sf]{subfig}
\usepackage{textcomp}
\usepackage{stfloats}
\usepackage{url}
\usepackage{verbatim}
\usepackage{graphicx}
\usepackage{cite}
\usepackage{tikz}
\usepackage{pgfplots}
\usepackage[hidelinks]{hyperref}

\begin{document}

\title{User-Feedback-Driven Adaptation for Vision-and-Language Navigation}


\author{%
  Yongqiang~Yu,~Xuhui~Li,~Hazza~Mahmood,~Jinxing~Zhou,~Haodong~Hong, \\~Longtao~Jiang,~Zhiqiang~Xu,~Qi~Wu,~and~Xiaojun~Chang%
  \thanks{Yongqiang Yu, Xuhui Li, Hazza Mahmood, Jinxing Zhou, Zhiqiang Xu, and Xiaojun Chang are with the Mohamed bin Zayed University of Artificial Intelligence (MBZUAI), Abu Dhabi, United Arab Emirates (e-mail: \{yongqiang.yu, xuhui.li, hazza.mahmood, jinxing.zhou, zhiqiang.xu, xiaojun.chang\}@mbzuai.ac.ae).}%
  \thanks{Haodong Hong is with the School of Electrical Engineering and Computer Science, The University of Queensland, Australia (e-mail: haodong.hong@uq.edu.au).}%
  \thanks{Longtao Jiang is with the University of Science and Technology of China (USTC), Hefei, China (e-mail: longtao.jiang@ustc.edu.cn).}%
  \thanks{Qi Wu is with the School of Computer Science, The University of Adelaide, Australia (e-mail: qi.wu01@adelaide.edu.au).}%
  \thanks{Xiaojun Chang is also with the University of Science and Technology of China (USTC), Hefei, China (e-mail: cxj@ustc.edu.cn).}%
  \thanks{Corresponding authors: Qi Wu (e-mail: qi.wu01@adelaide.edu.au) and Xiaojun Chang (e-mail: xiaojun.chang@mbzuai.ac.ae).}%
}



\maketitle

\begin{abstract}

Real-world deployment of Vision-and-Language Navigation (VLN) agents is constrained by the scarcity of reliable supervision after offline training. While recent adaptation methods attempt to mitigate distribution shifts via environment-driven self-supervision (e.g., entropy minimization), these signals are often noisy and can make the agent amplify its own mistakes during long-horizon sequential decision-making. In this paper, we propose a paradigm shift that positions user feedback, specifically episode-level success confirmations and goal-level corrections, as a primary, general-purpose supervision signal for VLN. Unlike internal confidence scores, user feedback is intent-aligned and in-situ consistent, directly correcting the agent's decoupling from user instructions. To effectively leverage this supervision, we introduce a user-feedback-driven learning framework featuring a topology-aware trajectory construction pipeline. This mechanism ``lifts'' sparse, goal-level corrections into dense path-level supervision. It does so by generating feasible paths on the agent’s incrementally built topological graph, enabling sample-efficient imitation learning without requiring step-by-step human demonstrations. Furthermore, we develop a persistent memory-bank mechanism for warm-start initialization, supporting the reuse of previously acquired topology and cached representations across navigation sessions. Extensive experiments on the GSA-R2R benchmark demonstrate that our approach transforms sparse interaction into robust supervision, consistently outperforming environment-driven baselines while exhibiting strong adaptability across diverse instruction styles. Our code is available at \url{https://github.com/seaotter999/UFD}.

\end{abstract}


\section{Introduction}

\IEEEPARstart{V}{ision-and-Language} Navigation (VLN)~\cite{Anderson_2018} aims to endow embodied agents with the ability to follow natural-language instructions and perform perception, understanding, and decision-making in 3D environments\cite{Chang_2017_Matterport3D,Savva_2019_Habitat}. Despite strong performance on standard benchmarks~\cite{Anderson_2018,Qi_2020,thomason2020vision,krantz2020beyond}, real-world deployment inevitably exposes agents to out-of-distribution perceptual\cite{Wang_2025_ICCV} and interactive conditions\cite{Li_2024_NeurIPS_HAVLN}. Specifically, unseen environments challenge model generalization, the high visual similarity of indoor scenes increases localization uncertainty, and variations in user referring habits and linguistic styles introduce significant semantic ambiguity. Such deployment-time distribution shift makes agents prone to drifting off course in the early stages of navigation, which in turn causes the observation stream to gradually decouple from the intended instruction. The resulting mismatch is further amplified through long-horizon sequential decision-making, leading to compounding errors\cite{Wang_2022_cotta}.

\begin{figure*}[!t]
\centering
\includegraphics[width=\textwidth]{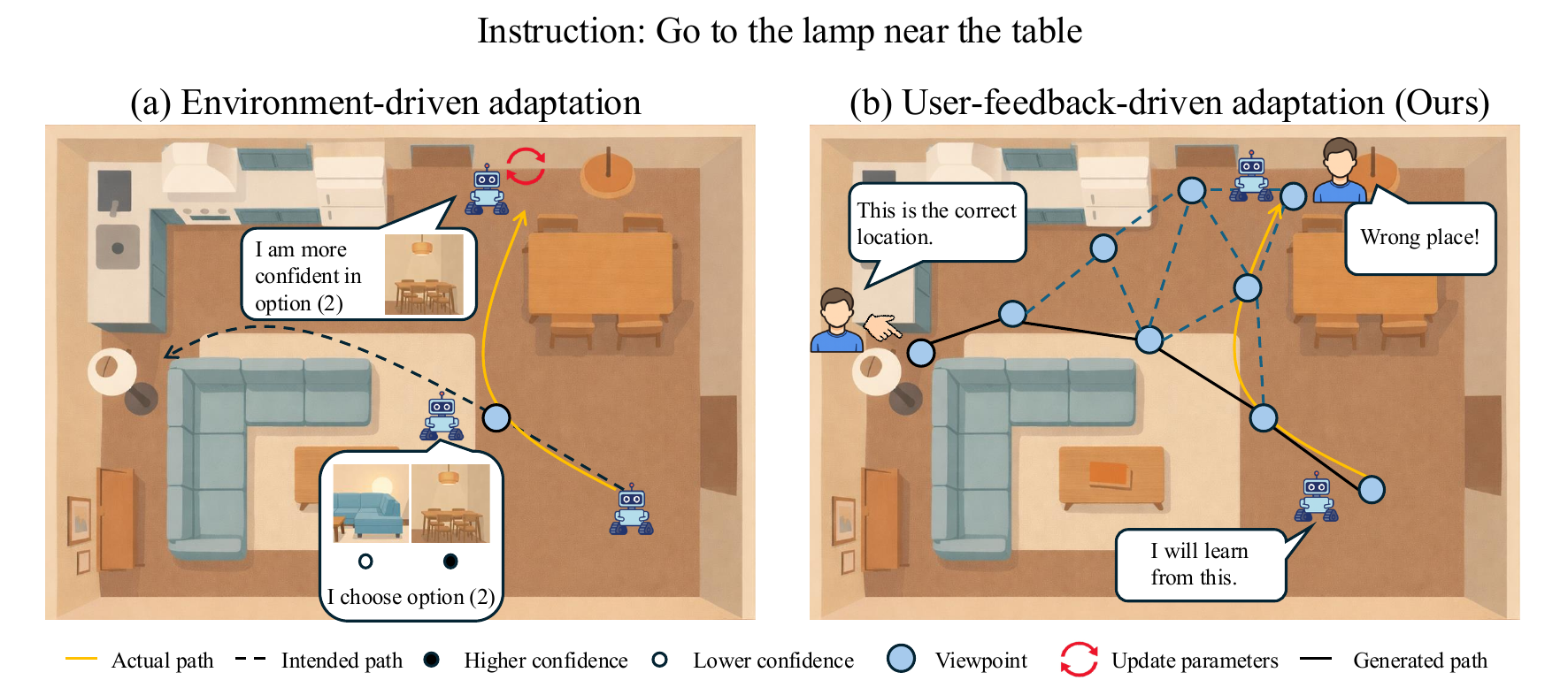}
\caption{(a) Environment-driven adaptation: Existing methods rely on self-supervision signals (e.g., entropy minimization), which are prone to error self-reinforcement when the agent drifts off the intended path. (b) User-feedback-driven adaptation (Ours): Our framework utilizes episode-level user feedback (success confirmation or goal-level correction). By leveraging the agent's constructed topological graph, we ``lift'' this sparse, goal-level correction into dense, path-level supervision. This provides intent-aligned guidance that effectively corrects the agent’s decoupling from instructions, preventing the accumulation of errors inherent in unsupervised or self-supervised methods.}
\label{fig_overview}
\end{figure*}

However, existing VLN evaluation\cite{jain2019stay} and adaptation paradigms still lack a systematic solution for obtaining reliable external supervision during deployment. Classical zero-shot evaluation\cite{ku2020RxR} typically freezes model parameters after offline training and performs one-shot testing in unseen environments\cite{zhang2025flexvln}. More recent unsupervised\cite{Wang_TENT_2021,Gao_FSTTA_2024} or self-supervised\cite{ttt,lu2022anticipating} test-time adaptation (TTA) methods allow updates at deployment time, as shown in Fig.~\ref{fig_overview}(a), yet their learning signals are largely derived from the agent’s own interaction experience, such as trajectory-based cues, confidence signals, or entropy-minimization objectives. In this paper, we refer to this family of approaches as environment-driven adaptation. For sequential decision-making tasks, such signals are often sparse and highly vulnerable to error contamination: early deviations alter subsequent observation distributions and cause mistakes to propagate along the trajectory; updating the model with pseudo-supervision generated from these deviated trajectories can further induce error self-reinforcement\cite{Ko_ATENA_2025}. Beyond environment-driven self-supervision, recent VLN TTA methods\cite{Ko_ATENA_2025, Kim_et_al_2025} have also explored incorporating external feedback; however, their supervision is typically coarse (success/failure) or selectively triggered, which remains insufficient to provide grounded, spatially precise corrections for long-horizon drift.
In other words, current deployment-time adaptation paradigms still struggle to obtain stable and reliable supervision in the long run.

To address this limitation, we view user feedback as a general form of supervision for VLN deployment.
Concretely, as depicted in Fig.~\ref{fig_overview}(b), this user feedback consists of episode-level success/failure confirmation after each navigation attempt, together with a goal-level correction upon failure—abstracted in our simulation setting as the correct target viewpoint—thereby providing intent-aligned external supervision.
This feedback has three key properties. First, it is intent-aligned\cite{Losey2021PhysicalInteraction}, directly reflecting the user’s true intended destination. Second, it is in-situ consistent, matching the topology and semantic conventions of the deployed environment while naturally capturing the user’s expression habits. Third, it is interaction-sparse, providing supervision at the episode level rather than step-by-step action demonstrations. 
Crucially, this signal serves as a unified correction mechanism. 
It naturally resolves both visual uncertainties and linguistic ambiguities by directly providing the intended target, regardless of the specific error source.
In this sense, user feedback not only helps the agent correct mistakes during deployment, but also turns deployment interactions into accumulative, trainable supervisory data.

Building on this perspective, we propose a user-feedback-driven learning framework. During deployment, the agent incrementally constructs a topological and memory-based representation of the environment. After each navigation episode, it receives sparse success/failure feedback and, when necessary, a goal-level correction. The system then converts such corrections into trajectory-level supervision, elevating the correction signal into high-quality training samples for imitation learning. Crucially, we do not require users to provide step-by-step guidance. 
Instead, we leverage the constructed topology to extrapolate path-level supervision from goal-level corrections. By computing a feasible path to the corrected goal on the topological graph, we effectively transform sparse feedback into dense supervision.
Furthermore, when deployment involves cross-session operation or redeployment after model updates, we introduce a persistent and reloadable memory-bank warm-start mechanism. This allows the system to reuse previously acquired scene topology and cached perceptual representations, mitigating early-stage performance degradation after updates.

To systematically evaluate the benefit of user-feedback supervision under controlled conditions, we conduct experiments on the GSA-R2R\cite{Hong_et_al_2025} benchmark. This benchmark provides richer instruction–trajectory samples within the same scenes and includes multiple instruction styles, such as Basic and the more realistic User style, enabling quantitative analysis of in-scene adaptation from deployment interactions. In implementation, we build on a graph-memory navigation agent, GR-DUET, since its explicit topology naturally supports graph-based correction mechanisms that lift goal-level feedback into trajectory-level supervision.

We further study two practical feedback streams. Continual adaptation updates the agent sequentially as feedback arrives over time from different users. Hybrid adaptation aggregates feedback collected within a deployment window and performs a single update to reduce update frequency and training cost. We evaluate both settings to demonstrate robustness under different deployment logistics.

Our main contributions are summarized as follows:

\begin{itemize}
    \item \textbf{User-Feedback-Driven Adaptation Paradigm}: We introduce a novel deployment-time adaptation paradigm that shifts reliance from unstable environment-driven self-supervision to intent-aligned user feedback. By leveraging episode-level success confirmations and goal-level corrections, our approach effectively mitigates error self-reinforcement inherent in existing entropy-based methods.

    \item \textbf{Topology-Aware Trajectory Construction}: We propose a topology-aware trajectory construction pipeline that transforms sparse, goal-level corrections into dense, path-level supervision. Utilizing the agent's incrementally built topological graph, this mechanism enables sample-efficient imitation learning without requiring labor-intensive, step-by-step human demonstrations.

    \item \textbf{Persistent Memory-Bank Warm-Start}: We develop a reloadable, persistent memory-bank mechanism designed for cross-session redeployment. This allows the agent to retain and reuse learned structural and perceptual knowledge across sessions, mitigating cold-start performance degradation.
    
    \item \textbf{Flexible Adaptation Strategies \& Validation}: We validate the framework's robustness by implementing two distinct adaptation modes: Continual Adaptation for sequential updates and Hybrid Adaptation for aggregating feedback from multiple users over a deployment window. Extensive experiments on the GSA-R2R benchmark demonstrate that our framework consistently outperforms environment-driven baselines on the standard Basic split, while exhibiting robust adaptability across diverse instruction styles in realistic deployment scenarios.

\end{itemize}

The rest of this paper is organized as follows. Section \ref{sec:related_work} reviews related work on VLN paradigms, test-time adaptation, and feedback-driven learning. Section \ref{sec:method} details our proposed user-feedback-driven adaptation framework. Section \ref{sec:experiments} presents experimental results and comprehensive analysis on the GSA-R2R benchmark. Finally, Section \ref{sec:conclusion} concludes the paper.

\section{Related Work}
\label{sec:related_work}

\subsection{Vision-and-Language Navigation Paradigms}
Vision-and-Language Navigation (VLN) has evolved significantly from early LSTM-based baselines\cite{Anderson_2018} to powerful Transformer-based architectures. Models such as Recurrent VLN BERT\cite{hong2021vln}, HAMT\cite{Chen_HAMT_2021}, and DUET\cite{Chen_DUET} capture long-range cross-modal dependencies, establishing strong performance on standard benchmarks. 
Parallel to these architectural advancements, scaling up supervision using web data has been a key strategy. Early efforts like Majumdar et al.\cite{Majumdar_2020} and AirBERT\cite{Guhur_Airbert_2021} utilized image-text pairs scraped from the web (e.g., Airbnb) to pre-train visual-linguistic representations. More recently, subsequent studies\cite{ytvln} extended this direction by learning navigation behaviors directly from YouTube videos, effectively utilizing open-world data to reduce dependency on environment-specific annotations.

Since the introduction of R2R, VLN has been predominantly evaluated under an unseen-environment protocol (val-unseen/test-unseen), i.e., agents are trained on one set of buildings and tested on distinct unseen buildings directly, with fixed parameters.
More recently, foundation-model-based approaches have pushed toward a stricter form of “zero-shot” VLN, aiming to navigate with minimal or no task-specific fine-tuning by leveraging VLM/LLM-based reasoning and planning. For instance, methods focusing on continuous environments\cite{chen2025constraint} introduce constraint-aware mechanisms to guide navigation without prior demonstrations. Concurrently, other studies\cite{qiao2025open} exploit the reasoning capabilities of Large Language Models (LLMs) to decompose high-level instructions into executable actions, enabling agents to navigate unseen environments by leveraging the broad knowledge encoded in foundation models.

While end-to-end learning remains dominant, modular architectures have gained traction for their interpretability and robustness. Specialized frameworks\cite{zhou2023esc,yitzhak2022cows} decouple exploration from object localization to handle object-goal navigation more effectively, while map-learning techniques\cite{chen2022weakly} have been proposed to aid spatial decision-making through weak supervision. However, despite these advancements in offline capabilities, these paradigms typically freeze model parameters during deployment. They lack the mechanism to adapt to dynamic distribution shifts or specific user preferences in real-time, highlighting the critical need for test-time adaptation.

\subsection{Test-Time Adaptation for VLN}

Test-Time Adaptation (TTA) has emerged as a promising direction to adapt pre-trained models to unseen distributions during deployment. In standard computer vision tasks, approaches like TENT\cite{Wang_TENT_2021} effectively adapt models by minimizing prediction entropy on test data. Inspired by this, recent works have extended TTA to the sequential nature of VLN. Notably, the FSTTA framework\cite{Gao_FSTTA_2024} introduces a fast-slow adaptation mechanism that balances structural stability with plasticity, allowing agents to adjust to environmental variations without catastrophic forgetting.

Beyond basic entropy minimization, other methods explore diverse supervision signals to enhance adaptation. For instance, ATENA\cite{Ko_ATENA_2025} incorporates a mixture entropy optimization strategy to mitigate overconfidence in incorrect actions and utilizes a self-active learning module to request feedback under high uncertainty. Similarly, FeedTTA\cite{Kim_et_al_2025} formulates adaptation as a reinforcement learning problem using binary success signals, while the Elastic Adaptation Model (EAM)\cite{eam} leverages a sample replay mechanism to adapt pre-trained agents in a source-free setting.

However, these approaches face critical limitations in providing effective supervision. Self-supervised methods (e.g., TENT, FSTTA) rely on intrinsic signals like entropy, which can lead to error self-reinforcement when the agent is overconfident in incorrect actions\cite{ross2011reduction,bengio2015scheduled}. On the other hand, while recent feedback-aware methods (e.g., FeedTTA, ATENA) incorporate external signals, they typically rely on sparse binary indicators (success/failure)\cite{Kim_et_al_2025} or internal uncertainty estimates to trigger feedback\cite{Ko_ATENA_2025}. Such signals often lack the dense, spatial granularity required to correct complex sequential deviations. This highlights the critical need for a framework that converts sparse interactions into dense, trajectory-level supervision, effectively halting the accumulation of errors.

\subsection{Feedback-Driven Learning and Continual Adaptation}

Reliable deployment requires agents to specialize in persistent environments over time. The General Scene Adaptation (GSA) paradigm\cite{Hong_et_al_2025} addresses this need by enabling agents to accumulate knowledge within specific scenes through repeated execution. However, standard GSA relies predominantly on unsupervised adaptation from self-exploration. While other approaches attempt to improve adaptability by generating synthetic instructions\cite{Wang_ScaleVLN_2023} or augmenting supervision with web-based image-text pairs\cite{Majumdar_2020, Guhur_Airbert_2021}, these offline methods often fail to capture the specific linguistic habits and diverse referring expressions produced by real users during actual deployment\cite{jain2019stay, wei2025unseen}.

Recent advances, such as SE-VLN\cite{dong2025se}, have explored self-evolution by utilizing Multimodal Large Language Models (MLLMs) for self-reflection. Yet, relying solely on internal model loops lacks the grounded correction that only a human can provide. Our work bridges this gap by introducing a user-feedback-driven framework. Unlike prior methods that rely on coarse success signals or unstable self-supervision, we leverage explicit user feedback, specifically success confirmations and goal-level corrections, as a reliable external supervision source. By systematically converting these sparse interactions into dense, trajectory-level training signals and employing a memory-bank warm-start, our approach enables robust, continual adaptation tailored to the specific user and environment.

\section{Method}
\label{sec:method}
\subsection{Task Definition: VLN and GSA-VLN}

In the standard Vision-and-Language Navigation (VLN) task, an embodied agent executes a natural language instruction in a novel environment without prior exposure.  
Formally, an instruction is represented as
\begin{equation}
I = (w_1, w_2, \dots, w_L),
\end{equation}
and the environment is modeled as a connectivity graph $G_{env}=(V,E)$, where $V$ is the set of discrete viewpoints and $E$ is the set of traversable edges connecting neighboring viewpoints. 
The agent begins from an initial viewpoint $v_1$, receives observation $o_t$ at step $t$, and selects an action:
\begin{equation}
a_t = \pi_\theta(I, v_t, o_t),
\end{equation}
where $\pi_\theta$ is a navigation policy parameterized by $\theta$.  

At each step $t$, the agent selects a discrete navigation action from the navigable neighbors of $v_t$ plus a \texttt{STOP} action. Specifically, we define the set of neighbors as $\mathcal{N}(v_t)=\{v' \in V \mid (v_t,v') \in E\}$, and the action space as $\mathcal{A}(v_t)=\mathcal{N}(v_t)\cup\{\texttt{STOP}\}$.

This process generates a trajectory $\tau=(v_{1},v_{2},...,v_{T})$, where each $v_t \in V$ represents a discrete viewpoint in the connectivity graph. The trajectory ideally terminates at the target viewpoint $v^{*}$ specified by $I$.
In this conventional formulation, model parameters are fixed during inference, and each episode is independent.

\paragraph{General Scene Adaptation (GSA-VLN)}
The GSA-VLN paradigm extends the VLN setting to more realistic, long-term deployments in which the agent repeatedly operates within the same environment.  
Instead of discarding prior experience, the agent maintains an environment-specific memory that accumulates structural and behavioral knowledge over time:
\begin{equation}
\mathcal{M} = \{(I_i, O_i, A_i, P_i)\}_{i=1}^{k},
\end{equation}
where $I_i$, $O_i$, $A_i$, and $P_i$ denote instructions, observations, actions, and paths collected from previous episodes.  
During subsequent runs, the agent can exploit this memory to inform navigation:
\begin{equation}
a_t = \pi_\theta(I, v_t, o_t, \mathcal{M}),
\end{equation}
and optionally update its parameters through unsupervised learning:

\begin{equation}
    \theta^{\prime} \leftarrow \text{Optimize}(\theta, \nabla_{\theta}\mathcal{L}(\mathcal{M};\theta)),
    \label{eq:update_unsup}
\end{equation}

where $\text{Optimize}(\cdot)$ denotes the parameter update step using a specific optimizer (e.g., AdamW), and $\mathcal{L}$ represents a self-supervised objective computed on stored experience.

\paragraph{Task Definition: User-Feedback-Driven Adaptation}
\label{sec:task_definition}

While standard GSA-VLN relies on unsupervised exploration, our framework incorporates explicit human guidance. We formulate the adaptation task through the following two definitions:

\noindent \textbf{Feedback Triggering Condition.}
Unlike step-by-step supervision, we adopt a realistic, interaction-sparse setting triggered at the \textit{episode level}. 
Let $T_{max}$ be the maximum allowed trajectory length. The episode terminates at time step $T$, defined as:
\begin{equation}
    T = \min \{t \mid a_t = \text{\texttt{STOP}} \lor t = T_{max} \}.
    \label{eq:T}
\end{equation}
The feedback is triggered if and only if the current step is the terminal step. Mathematically, this is expressed by the indicator function $\text{Tr}(t)$:
\begin{equation}
    \text{Tr}(t) = \mathbb{I} (t = T).
    \label{eq:trigger}
\end{equation}

\noindent \textbf{Feedback Formalism.}
The user provides a sparse supervision signal $\mathcal{F}$, which we define as a conditional correction function:
\begin{equation}
\mathcal{F}(v_T, v^{*})=
\begin{cases}
v^{*}, & \text{if } v_T\neq v^{*}\ \text{(Correct Goal)},\\
v_T, & \text{if } v_T = v^{*}\ \text{(Confirmed Stop)}.
\end{cases}
\label{eq:feedback_formalism}
\end{equation}

When the agent fails ($v_T \neq v^*$), the function returns the correct target $v^*$. 
When the agent succeeds, the user provides a success confirmation, and we treat the terminal viewpoint $v_T$ as the feedback-validated endpoint. 
This feedback-validated endpoint is utilized to derive a supervisory trajectory $\tau^+$ by leveraging the environment connectivity $G$ (e.g., via A* search). 
The resulting instruction--trajectory pairs are accumulated into an adaptation dataset 
$\mathcal{D}_{\mathrm{adapt}} = \{(I_j, \tau^+_j)\}_{j=1}^{N}$, where $N$ denotes the number of constructed adaptation samples.

Finally, unlike the unsupervised update in Eq.~\ref{eq:update_unsup}, the policy is refined by minimizing a feedback-driven imitation objective:
\begin{equation}
    \theta^{\prime} \leftarrow \text{Optimize}(\theta, \nabla_{\theta}\mathcal{L}_{feedback}(\mathcal{D}_{adapt};\theta)),
    \label{eq:update_feedback}
\end{equation}
where $\mathcal{L}_{feedback}$ measures the discrepancy between the agent's policy and the corrected trajectories in $\mathcal{D}_{adapt}$.

\newcommand{\icondash}{
  \raisebox{-0.3\height}{\includegraphics[width=1em]{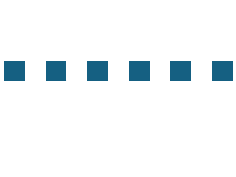}}
}
\newcommand{\icondouble}{
  \raisebox{-0.3\height}{\includegraphics[width=1em]{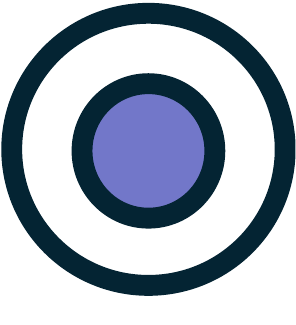}}
}
\newcommand{\icondarkblue}{
  \raisebox{-0.3\height}{\includegraphics[width=1em]{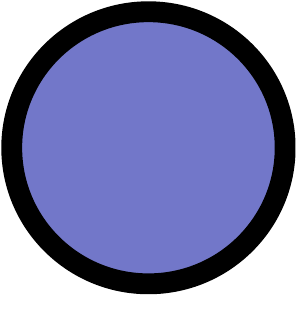}}
}
\begin{figure*}[!t]
\centering
\includegraphics[width=\textwidth]{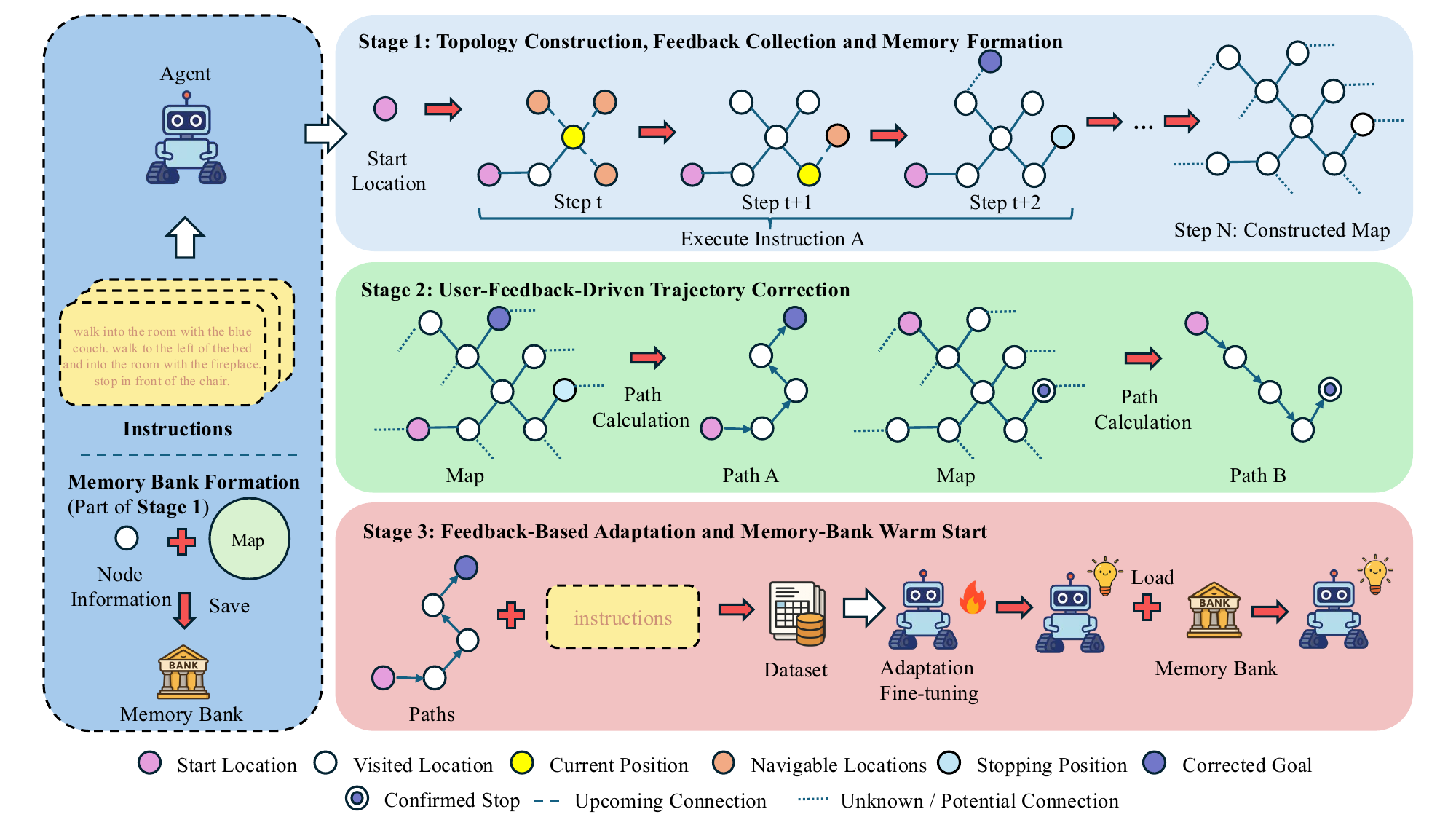}
\caption{
Overview of the proposed user-feedback-driven adaptation framework for Vision-and-Language Navigation (VLN). 
\textbf{Stage 1:  Topology Construction, Feedback Collection and Memory Formation.} The agent executes instructions while incrementally constructing a topological graph and caching node information into a persistent memory bank.
The dashed lines\protect\icondash represent unknown (e.g., Step t+2) or potential (e.g., constructed map) connections.
At episode termination, user feedback is collected as either a goal-level correction (Correct Goal \protect\icondarkblue) or a success confirmation (Confirmed Stop \protect\icondouble).
\textbf{Stage 2: User-Feedback-Driven Trajectory Correction.} 
The agent derives trajectory-level training signals from goal-level corrections. Using A* search on the topological graph, it computes feasible paths toward the Corrected Goal or Confirmed Stop.
\textbf{Stage 3: Feedback-Based Adaptation and Memory-Bank Warm Start.} The corrected instruction-trajectory pairs are aggregated to fine-tune the policy using the DAgger framework. During memory-bank warm start, the agent reloads the memory bank, which stores the environment topology and cached node information, to enable warm-start initialization and mitigate cold-start degradation.
}
\label{pipeline}
\end{figure*}

\subsection{Stage 1: Topology Construction, Feedback Collection and Memory Formation}

As illustrated in Stage 1 of Fig.~\ref{pipeline}, the agent starts with no prior knowledge of the environment. During navigation, it dynamically builds a topological graph $G=(V, E)$ by identifying navigable viewpoints and their connectivity.
Specifically, at each step $t$, the agent visits the current viewpoint $v_t$ and identifies its navigable neighbors. If $v_t$ or any neighbor is new, they are added to the node set $V$, and the spatial connections are recorded in $E$. This process incrementally expands the map from a single starting point to a connected graph covering the visited areas. Following GR-DUET, we maintain this structure in an environment-level memory $\mathcal{M}^{\prime}=(G, \mathcal{C}, \mathcal{S})$, but extend it to be persistently reloadable. $G$ stores the discovered topology, $\mathcal{C}$ caches the panoramic visual features and positional encodings of visited nodes to preserve a persistent environmental representation across sessions, and $\mathcal{S}$ records candidate-viewpoint mappings. This constructed memory serves as the foundation for both trajectory correction (Stage 2) and memory-bank warm-start initialization (Stage 3).

In addition to map building, this stage executes the critical feedback collection process. When an episode terminates at $v_T$ (Eq.~\ref{eq:T}), we query the user for episode-level feedback (Eq.~\ref{eq:feedback_formalism}).
The feedback either confirms a correct stop (Confirmed Stop) or provides a corrected goal viewpoint $v^*$ upon failure (Correct Goal).
These collected feedback signals, together with the constructed memory $M^{\prime}$, are then used in Stage 2 to compute corrected trajectories and lift sparse goal-level feedback into dense trajectory-level supervision.

\subsection{Stage 2: User-Feedback-Driven Trajectory Correction}

Following the episode-level triggering condition in Sec.~III-A (Eq.~\ref{eq:T}--\ref{eq:trigger}),
the agent executes an instruction $I$ until termination at $T$, producing a trajectory
\begin{equation}
    \tau = (v_1, v_2, \dots, v_{T}),
\end{equation}
with terminal viewpoint $v_{T}$.
User feedback is queried only at this terminal step (i.e., $\mathrm{Tr}(T)=1$).
Let $\hat{v} = \mathcal{F}(v_T, v^{*})$ denote the supervision endpoint (Sec. III-A, Eq.~\ref{eq:feedback_formalism}),
which is either the corrected goal $v^{*}$ (upon failure) or the confirmed stopping viewpoint $v_T$ (upon success).
We then convert this sparse feedback to trajectory-level supervision by computing a feasible path on the current topological graph $G$:
\begin{equation}
    \tau^{+} = A^{*}(v_1 \to \hat{v}, G).
\end{equation}

Each corrected pair $(I, \tau^{+})$ is stored as a new adaptation sample.

To ensure data quality, we apply a length-based outlier rejection on the corrected trajectory $\tau^+$. Let $\lvert\tau^+\rvert$ denote the number of nodes (viewpoints) in $\tau^+$.
Importantly, the thresholds are \emph{not} chosen heuristically or tuned on downstream performance; instead, they are determined directly from the trajectory-length distribution of our corrected paths.
Across 12{,}000 corrected trajectories constructed in Stage~2, the path length ranges from 1 to 14 with an average of 6.01.
The distribution is highly concentrated: 94.5\% of trajectories fall within 5--7 nodes (length 5: 24.2\%, length 6: 34.8\%, length 7: 35.5\%).
Therefore, we keep only trajectories with $5 \le \lvert\tau^+\rvert \le 7$ (discarding $\lvert\tau^+\rvert<5$ or $\lvert\tau^+\rvert>7$),
removing only 5.5\% outliers, which are often caused by incomplete topology or corrections inconsistent with the discovered graph.

\subsection{Stage 3: Feedback-Based Adaptation and Memory-Bank Warm Start}

Once corrected trajectories are accumulated, we form an adaptation dataset:
\begin{equation}
\mathcal{D}_{adapt} = \{(I_i, \tau_i^+)\}_{i=1}^{N}.
\end{equation}
The navigation policy is then fine-tuned using the standard imitation learning objective adopted in GR-DUET:
\begin{equation}
\label{eq:il}
\mathcal{L}_{IL} = - \frac{1}{N} \sum_{(I, \tau^+) \in \mathcal{D}_{adapt}}
\sum_{t=1}^{|\tau^+|} \log \pi_\theta(a_t^+ \mid v_t, I),
\end{equation}
where $a_t^+$ is the oracle action from the corrected trajectory $\tau^+$.  
This objective aligns the policy with user feedback without modifying the base model’s optimization procedure.

\subsubsection{Memory-bank Warm Start}

After fine-tuning, the agent is redeployed in the same environment.  
Instead of reconstructing the graph and visual features from scratch, it reloads the previously stored memory $\mathcal{M}^{\prime}=(G, \mathcal{C}, \mathcal{S})$, enabling immediate reuse of environment topology and cached representations.  
This \emph{reloadable memory} extends the GR-DUET design by introducing persistent scene states that can be saved and restored across sessions, maintaining structural and perceptual continuity during redeployment.

Although the memory-bank warm start does not alter the training objective, it provides a practical advantage for long-term operation.  
By retaining environment-specific knowledge acquired in previous runs, it stabilizes early-stage navigation and mitigates cold-start degradation that often occurs after model updates.  
This persistent memory thus serves as a lightweight yet effective mechanism to enhance robustness and ensure smooth adaptation across continual and hybrid deployment settings.

\subsubsection{Continual Adaptation}

In long-term deployment, an agent interacts with multiple users sequentially over time.  
To enable continuous improvement, we perform incremental parameter updates as new feedback data $\mathcal{D}_{adapt}^{(t)}$ are collected at time $t$:
\begin{equation}
\theta_{t+1} = \theta_t - \alpha \nabla_\theta \mathcal{L}_{IL}(\theta_t;\mathcal{D}_{adapt}^{(t)}),
\label{eq:continual}
\end{equation}
where $\alpha$ is the learning rate.  
This continual adaptation mechanism allows the agent to gradually integrate feedback from successive interactions, aligning its behavior with both evolving user language and environment familiarity.

\subsubsection{Hybrid Adaptation}

In practical deployments such as households, the same agent may serve multiple users in the same environment, and their feedback may accumulate within a deployment window.
To handle such aggregated supervision efficiently, we introduce \emph{hybrid adaptation}, which aggregates feedback from all users within a deployment window:
\begin{equation}
\mathcal{D}_{\text{hybrid}} = \bigcup_{u=1}^{U} \mathcal{D}_u, \qquad
\theta' = \arg\min_\theta \mathcal{L}_{IL}(\theta;\mathcal{D}_{\text{hybrid}}),
\label{eq:hybrid}
\end{equation}
where $\mathcal{D}_u$ represents feedback-corrected trajectories from user $u$.  
By jointly optimizing across multiple user datasets, hybrid adaptation captures diverse linguistic expressions and navigation styles while avoiding bias toward early interactions.

Consequently, the proposed framework supports two distinct update modes: sequential updates for continuous adaptation (Eq.~\ref{eq:continual}) and batched aggregation for multi-user feedback (Eq.~\ref{eq:hybrid}). This flexibility allows the agent to handle varying streams of supervision signals consistent with real-world deployment logistics.

\section{Experiments}
\label{sec:experiments}
\subsection{Experimental Setup}

\paragraph{Datasets and Environments}
We conduct experiments on the GSA-R2R dataset~\cite{Hong_et_al_2025}, which extends the 
standard Room-to-Room (R2R) benchmark~\cite{Anderson_2018} to support scenario adaptation. 
In this work, we focus on the \emph{residential environments} and evaluate on two instruction 
styles: Basic and User. This setting captures the most common real-world 
deployment scenarios, where navigation agents repeatedly operate in household layouts and 
encounter both neutral and user-specific language patterns.

\paragraph{Baselines}
We adopt GR-DUET~\cite{Hong_et_al_2025} as our primary baseline, since it represents the 
state of the art for GSA-VLN by explicitly maintaining a global navigation graph for adaptation. To evaluate our method against existing adaptation strategies, we include representative deployment-time adaptation (TTA) baselines built upon GR-DUET, such as entropy-minimization methods (TENT, FSTTA) and the feedback-aware ATENA.
Finally, to further validate the generality of our approach, we also test other representative VLN methods under the GSA-VLN protocol, including DUET~\cite{Chen_DUET}, HAMT ~\cite{Chen_HAMT_2021}, and NaviLLM~\cite{Zheng_2024_CVPR}. These methods were originally 
developed for the standard zero-shot VLN setting, and their performance in the adaptation scenario serves as an additional reference.

\paragraph{Implementation Details}
Our method builds on GR-DUET as the backbone. We preserve its model architecture and loss 
formulation, and introduce our feedback-driven dataset construction and memory-bank warm start 
modules. We adopt the DAgger algorithm during training. At each iteration, 
the agent executes instructions using a mixture of its current policy 
and exploration sampling, while an expert oracle provides optimal 
actions for each encountered state. The expert policy can be configured 
to use either shortest path distances (SPL) or normalized Dynamic Time 
Warping scores (nDTW) to evaluate action quality relative to the ground 
truth trajectory. The policy is then updated using imitation learning 
on the aggregated state-action pairs from both expert demonstrations 
and student rollouts.

\paragraph{Adaptation Settings}
\label{subsec:adaptation_settings}
All adaptation experiments share consistent optimization and hardware settings. 
We optimize the model using the AdamW optimizer with a learning rate of $1\times 10^{-5}$ and a batch size of 2.
Each adaptation stage consists of 30,000 episodes.
Training is conducted on a single NVIDIA RTX~4090 GPU for about 82 hours per stage.

During adaptation, the agent is trained on a mixture of user-feedback data and source-domain data (i.e., R2R and augment dataset used by GR-DUET). 
This mixed-domain replay helps stabilize adaptation and mitigates overfitting to user-specific language, thereby maintaining general navigation capability while learning from user-feedback data.

We consider two adaptation modes under identical optimization settings but with different data scales:

\begin{itemize}
    \item \textbf{Continual adaptation.} The model undergoes three sequential user-guided updates (e.g., Basic $\rightarrow$ User~A $\rightarrow$ User~B), where each stage uses approximately 500 feedback samples. 
    Across all three stages, the cumulative exposure amounts to about 1,500 feedback samples, resulting in a proportionally longer total training time.
    \item \textbf{Hybrid adaptation.} The model aggregates feedback from five users into a single adaptation stage with 500 total samples. 
    This one-time adaptation corresponds to the same data scale as a single continual stage, requiring only about one-third of the total training time while providing a compact and efficient adaptation setup.
\end{itemize}

Unless otherwise specified, all subsequent adaptation experiments follow the above configuration.

\subsection{Evaluation Metrics}

We evaluate navigation performance using standard metrics from the Vision-and-Language Navigation (VLN) literature, covering goal achievement, path efficiency, trajectory fidelity, and semantic alignment. 
Following the Room-to-Room (R2R) benchmark~\cite{Anderson_2018}, we report nine widely used metrics: 
Success Rate (SR) \cite{Anderson_2018}, Oracle Success Rate (Oracle SR)\cite{Anderson_2018}, Success weighted by Path Length (SPL) \cite{anderson2018evaluation}, Navigation Error (NE) \cite{Anderson_2018}, Oracle Error / Oracle Navigation Error \cite{jain2019stay}, Path Length \cite{Anderson_2018}, Normalized Dynamic Time Warping (nDTW) \cite{ilharco2019general}, Success weighted by nDTW (SDTW) \cite{ilharco2019general}, and Coverage weighted by Length Score (CLS) \cite{jain2019stay}. 
These metrics are adopted consistently across recent VLN benchmarks~\cite{Anderson_2018,Ku_2020,jain2019stay,Hong_et_al_2025} to ensure comparability with prior work.

\subsection{Main Results on the GSA-Basic Test Set}

Table~\ref{tab:main_results} presents the quantitative comparison of different VLN methods on the GSA-Basic test split. 
We include representative baselines HAMT~\cite{Chen_HAMT_2021}, NaviLLM~\cite{Zheng_2024_CVPR}, DUET~\cite{Chen_DUET}, and GR-DUET~\cite{Hong_et_al_2025}, where GR-DUET serves as the baseline for our framework. 

We further include several deployment-time adaptation baselines on top of GR-DUET: entropy-minimization-based methods (TENT and FSTTA) and feedback-aware adaptation methods (ATENA). For all TTA baselines, we perform a hyperparameter sweep and report the best configuration under our evaluation protocol.
We also evaluate our method under two configurations: the feedback-driven model after adaptation training without memory loading (Ours w/o Mem) and the model evaluated with the reloadable memory bank (Ours w/ Mem).

\begin{table*}[!t]
\renewcommand{\arraystretch}{1.1}
\setlength{\tabcolsep}{8pt}
\caption{Performance comparison on the GSA-Basic test set. 
Higher SR, Oracle SR, SPL, nDTW, SDTW, and CLS indicate better results; lower NE, Oracle Error, and Path Length indicate better efficiency and precision. 
Best results are highlighted in \textbf{bold}.}
\label{tab:main_results}
\centering
\footnotesize
\begin{tabular}{lccccccccc}
\hline
Method & SR$\uparrow$ & Oracle SR$\uparrow$ & SPL$\uparrow$ & NE$\downarrow$ & Oracle Error$\downarrow$ & PL$\downarrow$ & nDTW$\uparrow$ & SDTW$\uparrow$ & CLS$\uparrow$ \\
\hline

HAMT ~\cite{Chen_HAMT_2021}&40.12  & 46.25 & 38.23 & 5.74 & 3.97 & \textbf{7.72} & 55.54 & 34.06 & 56.09 \\
NaviLLM ~\cite{Zheng_2024_CVPR}&49.92  & 60.25 & 43.23 & 4.65 & 3.01 &  10.25& 57.15 & 39.29 & 56.44 \\
DUET~\cite{Chen_DUET} & 54.04 & 65.46 & 42.26 & 4.44 & 2.69 & 13.53 & 53.06 & 38.71 & 52.13 \\
GR-DUET (baseline)~\cite{Hong_et_al_2025} & 66.96 & 76.58 & 61.22 & 3.25 & 2.02 & 9.80 & 69.24 & 56.38 & 67.18 \\
\hline

GR-DUET+TENT~\cite{Wang_TENT_2021}  & 67.46 & 76.96 & 61.65 & 3.23 & 2.01 & 9.69 & 69.47 & 56.84 & 67.30 \\

GR-DUET+FSTTA~\cite{Gao_FSTTA_2024} & 67.33 & 77.17 & 61.37 & 3.21 & 1.99 & 9.87 & 69.12 & 56.50 & 67.17 \\

GR-DUET+ATENA~\cite{Ko_ATENA_2025} & 65.69 & 72.76 & 61.89 & 3.38 & 2.34 & 8.52 & 69.97 &55.57 & 67.82\\
\hline
Ours (w/o Mem) & 76.33 & 83.21 & 70.99 & 2.53 & 1.67 & 9.03 & 77.13 & 66.65 & 73.55 \\
Ours (w/ Mem)  & \textbf{79.04} & \textbf{85.67} & \textbf{74.40} & \textbf{2.26} & \textbf{1.52} & 8.81 & \textbf{79.67} & \textbf{69.80} & \textbf{76.42} \\
\hline
\end{tabular}
\end{table*}

From Table~\ref{tab:main_results}, we observe that traditional transformer-based models such as DUET, NaviLLM, and HAMT exhibit limited performance on the GSA-Basic test split. 
These methods, originally optimized for static or single-pass VLN settings, struggle to adapt effectively to the long-term, environment-specific nature of GSA-VLN. 
In contrast, GR-DUET, which integrates graph-based memory for scene adaptation, provides a much stronger baseline and demonstrates clear advantages over conventional architectures.

Compared to GR-DUET, entropy-minimization-based TTA methods bring only marginal improvements (TENT: $+0.50$ SR / $+0.43$ SPL, FSTTA: $+0.37$ SR / $+0.15$ SPL), suggesting that self-supervised entropy signals are relatively weak in the GSA-VLN setting under distribution shift. ATENA presents a mixed trend. While it improves certain efficiency and fidelity metrics (e.g., SPL/nDTW/CLS) and achieves a shorter path length, it yields lower SR and Oracle SR in our setting. In contrast, our user-feedback-driven supervision provides consistent gains across almost all metrics.

\begin{figure*}[!tb]
\centering
\includegraphics[width=0.9\textwidth]{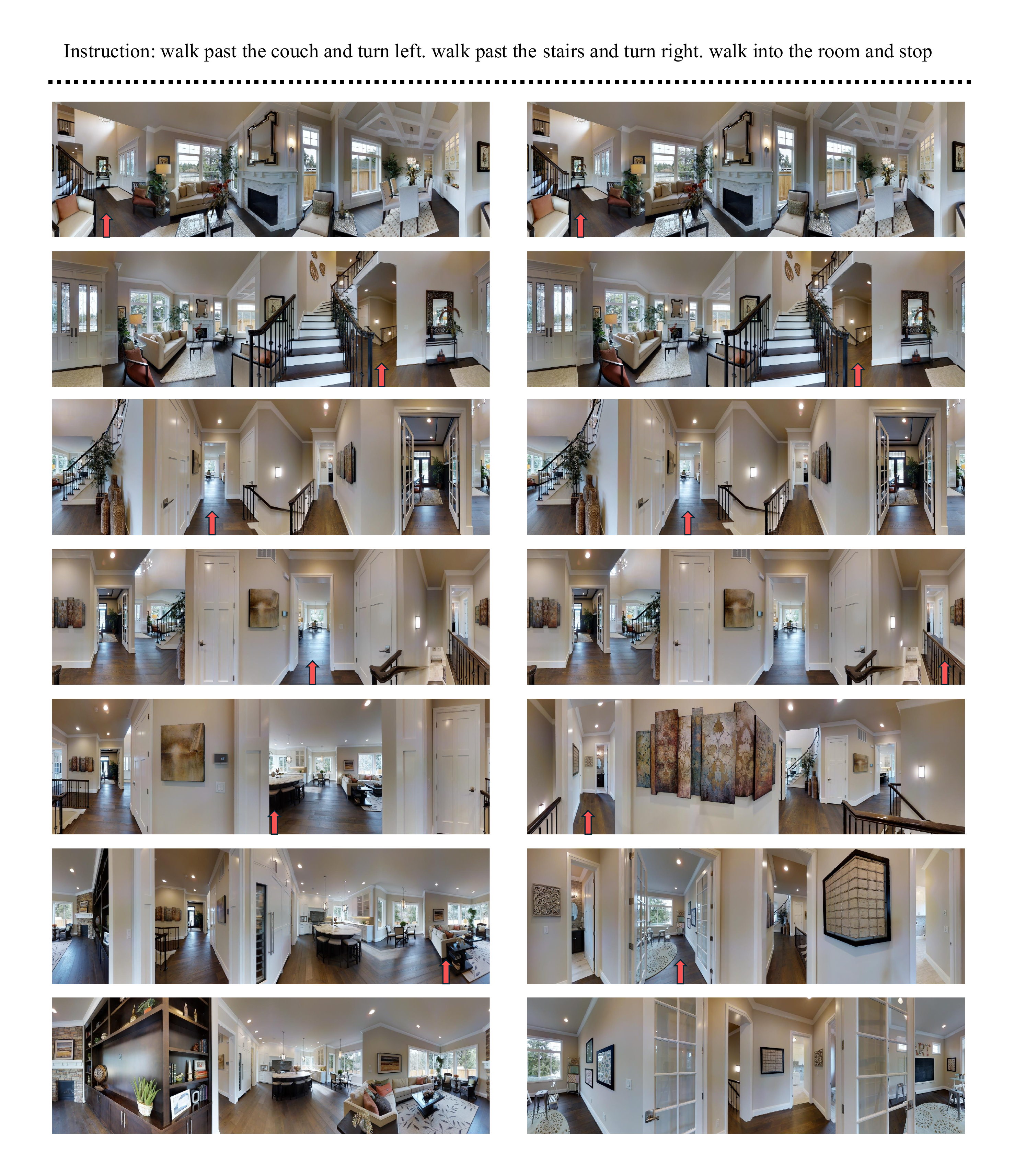}
\vspace{-4ex}
\caption{Comparison of GR-DUET navigation before (left) and after (right) user-feedback-driven adaptation, showing improved alignment between language instructions and executed trajectory.}
\label{fig_double}
\end{figure*}

Building upon GR-DUET, our proposed user-feedback-driven framework further enhances navigation performance across almost all evaluation metrics. 
The feedback-driven variant (Ours w/o Mem) achieves substantial improvement over GR-DUET, showing that incorporating user feedback significantly enhances both navigation success and path efficiency through explicit human-guided policy refinement. 
Enabling the reloadable memory bank (Ours w/ Mem) yields significant performance gains, demonstrating that previously explored environmental knowledge can be effectively leveraged to overcome the cold-start condition.
Without loaded memory, the model begins with limited environmental context, leading to temporarily suboptimal performance that improves as memory is reconstructed during navigation.
Overall, these findings suggest that user feedback acts as the primary driver 
of continual policy improvement, whereas the memory-bank warm-start mechanism 
offers complementary benefits by mitigating cold-start degradation across 
deployment cycles.
A qualitative visualization in Fig.~\ref{fig_double} further supports these results: compared to the baseline GR-DUET (left), our user-feedback-adapted model (right) more accurately executes the instruction “walk past the couch and turn left, walk past the stairs and turn right, walk into the room and stop,” successfully reaching the target room. This visual evidence aligns with the consistent improvements across all quantitative metrics in Table \ref{tab:main_results}.

\subsection{Ablation Study}

We further analyze how the amount of executed instructions during Stage~1 and the number of user feedback samples during Stage~2 jointly affect the adaptation performance. 
Table~\ref{tab:ablation_feedback} reports results under different combinations of instruction and feedback quantities on the GSA-Basic test split, using GR-DUET as the reference.

The results show a clear and consistent trend: both instruction execution and user feedback contribute positively to adaptation. 
Executing more instructions during environment exploration yields denser and more accurate topological graphs, thereby improving spatial understanding. 
Meanwhile, increasing the number of user feedback samples provides richer and higher-quality supervision, allowing the policy to better align with human intent. 
Notably, even with only 100 feedback samples, our model already surpasses the GR-DUET baseline by a large margin, demonstrating the high sample efficiency of feedback-driven learning. 
The best performance is achieved when both executed instructions and feedback samples are abundant (500/500), confirming that broader exploration and richer user corrections are complementary and jointly enhance agent adaptation.

\begin{table}[!t]
\renewcommand{\arraystretch}{1.2}
\setlength{\tabcolsep}{3pt}
\caption{Ablation results on the GSA-Basic set under different instruction and feedback quantities. The first row shows the \textbf{baseline} (default) configuration.}
\label{tab:ablation_feedback}
\centering
\footnotesize
\begin{tabular}{cc|ccccccc}
\hline
\#Instr. & \#Fb & SR$\uparrow$ & SPL$\uparrow$ & NE$\downarrow$ & PL$\downarrow$ & nDTW$\uparrow$ & SDTW$\uparrow$ & CLS$\uparrow$ \\
\hline
\textbf{500} & \textbf{500} & \textbf{76.33} & \textbf{70.99} & \textbf{2.53} & 9.03 & \textbf{77.13} & \textbf{66.65} & \textbf{73.55}\\
500 & 300 & 74.58 & 69.27 & 2.62 & 9.12 & 75.86 & 64.75 & 72.58 \\
500 & 100 & 73.83 & 67.59 & 2.68 & 9.26 & 75.05 & 63.57 & 71.48 \\
300 & 300 & 74.00 & 69.45 & 2.69 & \textbf{8.82} & 76.16 & 64.65 & 72.82 \\
300 & 100 & 73.38 & 68.19 & 2.80 & 8.93 & 74.92 & 63.77 & 71.60 \\
100 & 100 & 72.29 & 66.20 & 2.94 & 9.59 & 72.74 & 61.93 & 69.77 \\

\hline
\end{tabular}
\end{table}

We observe that environment exploration and user feedback play distinct but reinforcing roles. Increasing the number of executed instructions establishes a more comprehensive spatial representation, which in turn amplifies the effectiveness of semantic corrections provided by user feedback. Crucially, this combination enables robust adaptation even under limited supervision (e.g., 100 samples), demonstrating high sample efficiency.

\subsection{Effect of Executed Instruction Quantity}
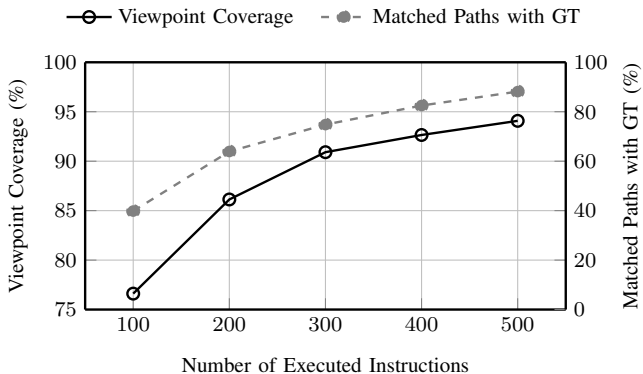
\begin{figure}[t]
\centering
\footnotesize
\begin{tikzpicture}

\begin{axis}[
    width=0.9\linewidth,
    height=0.55\linewidth,
    xlabel={Number of Executed Instructions},
    ylabel={Viewpoint Coverage (\%)},
    ylabel near ticks,
    ylabel style={font=\footnotesize},
    ymin=75, ymax=100, 
    ytick={75,80,85,90,95,100},
    xmin=50, xmax=550,
    xtick={100,200,300,400,500},
    axis y line*=left,
    axis x line*=bottom,
    axis line style={line width=0.9pt},
    every outer x axis line/.append style={line width=0.9pt},
    every outer y axis line/.append style={line width=0.9pt},
    grid=both,
    tick align=inside,
    line width=0.9pt, 
    mark size=2.2pt,
    legend columns=2,
    legend style={
        font=\footnotesize,
        at={(0.5,1.10)}, 
        anchor=south,
        draw=none, 
        fill=none,
        /tikz/every even column/.style={column sep=8pt}
    },
]

\addplot[color=black,mark=o,line width=0.9pt]
  coordinates {(100,76.61) (200,86.13) (300,90.91) (400,92.65) (500,94.08)};
\addlegendentry{Viewpoint Coverage}

\addlegendimage{color=gray,mark=*,dashed,line width=0.9pt}
\addlegendentry{Matched Paths with GT}

\end{axis}

\begin{axis}[
    width=0.9\linewidth,
    height=0.55\linewidth,
    xmin=50, xmax=550,
    ylabel={Matched Paths with GT (\%)},
    ylabel near ticks,
    ylabel style={font=\footnotesize},
    ymin=0, ymax=100, 
    ytick={0,20,40,60,80,100},
    axis y line*=right,
    axis x line*=top,
    xtick=\empty,
    grid=none,
    tick align=inside,
    line width=0.9pt, 
    mark size=2.2pt,
]

\addplot[color=gray,mark=*,dashed,line width=0.9pt]
  coordinates {(100,39.83) (200,63.92) (300,74.80) (400,82.51) (500,88.15)};

\end{axis}

\end{tikzpicture}
\caption{Effect of executed instruction quantity on both viewpoint coverage (left axis) and matched paths with ground-truth (right axis).}
\label{fig:dual_yaxis_coverage_match}
\end{figure}

To further analyze how environment exploration influences adaptation quality, we examine the relationship between the number of executed instructions and two key indicators: viewpoint coverage and matched paths with ground-truth trajectories, as shown in Fig.~\ref{fig:dual_yaxis_coverage_match}.
Viewpoint coverage measures the proportion of unique navigable viewpoints visited during navigation, reflecting the spatial completeness of the constructed topological graph.
Matched paths represent the percentage of reconstructed trajectories that correctly align with ground-truth routes. Each executed instruction is accompanied by one corresponding user feedback sample.
As the number of executed instructions increases from 100 to 500, both metrics exhibit consistent improvement.
Viewpoint coverage rises from 76.61\% to 94.08\%, indicating that broader exploration leads to more comprehensive spatial representations.
Similarly, the proportion of matched paths increases from 39.83\% to 88.15\%, confirming that richer environmental exposure leads to more accurate reconstruction of ground-truth routes.
These trends quantitatively support our claim that more extensive instruction execution enhances environment-specific representation and understanding, providing a stronger foundation for subsequent user-feedback-driven adaptation.

A concern in utilizing episode-level user feedback is whether accepting all trajectories that reach the correct target $v^*$, regardless of their intermediate path, introduces noise that hampers learning. To verify this, we conducted an additional analysis on the GSA-Basic split. We filtered the collected feedback samples to retain only those strictly matching the ground-truth shortest paths (reducing the dataset size from $\sim$11342 to 10,587 samples) and retrained the model. The results showed that the strictly filtered model achieved an SR of 76.33\% and SPL of 71.14\% at convergence, nearly identical to our proposed method (SR 76.33\%, SPL 70.99\%). This suggests that: (1) trajectories that successfully reach the target but deviate slightly from the oracle path still provide valid navigational signals; and (2) our feedback mechanism is robust to path variations, effectively leveraging episode-level supervision without requiring strict step-by-step alignment, which is often unavailable in real-world scenarios.

\begin{table}[!t]
\renewcommand{\arraystretch}{1.15}
\setlength{\tabcolsep}{3pt}
\caption{Continual adaptation over five runs.
Each run applies three sequential updates:
Adapt-1 (Basic) $\rightarrow$ Adapt-2/3 (randomly sampled from \{Child, Sheldon, Moira, Rachel, Keith\}).
We report Adapt-3 results and the gain over the \emph{best single-step} among Adapt-2/3 ($\Delta^\star$).}
\label{tab:continual_compact}
\centering

\resizebox{\columnwidth}{!}{
\begin{tabular}{c|l|cc|cc}
\hline
Run & Seq. (Adapt-2/3) & SR & $\Delta^\star$ & SPL & $\Delta^\star$ \\
\hline
1 & Basic$\to$Rachel$\to$Child    & 73.00 & $-0.71$ & 68.79 & $+0.16$ \\
2 & Basic$\to$Keith$\to$Rachel   & 74.29 & $+0.58$ & 69.19 & $+0.56$ \\
3 & Basic$\to$Moira$\to$Keith    & 72.48 & $+1.05$ & 67.64 & $+0.95$ \\
4 & Basic$\to$Sheldon$\to$Moira  & 72.52 & $+0.28$ & 68.85 & $+1.08$ \\
5 & Basic$\to$Child$\to$Sheldon  & 73.95 & $+1.71$ & 69.13 & $+1.36$ \\
\hline
Avg &  & \textbf{73.25} & \textbf{+0.58} & \textbf{68.72} & \textbf{+0.82} \\
\hline
\end{tabular}
}
\vspace{2pt}
\begin{minipage}{0.96\linewidth}
\footnotesize
\emph{Notes.} $\Delta^\star$: gain over the \emph{best} one-step fine-tuning among Adapt-2/3.
\end{minipage}
\end{table}

\subsection{Continual and Hybrid Adaptation Results}
\label{subsec:continual_hybrid}

\begin{table}[!t]
\setlength{\tabcolsep}{5pt}
\renewcommand{\arraystretch}{1.15}
\caption{Aggregate improvement of continual adaptation (Adapt-3) over the mean single-step fine-tuning across \{Child, Sheldon, Moira, Rachel, Keith\}.}
\label{tab:continual_aggregate}
\centering
\resizebox{\columnwidth}{!}{
\begin{tabular}{lcccccc}
\hline
 & SR$\uparrow$ & SPL$\uparrow$ & NE (m)$\downarrow$ & nDTW$\uparrow$ & SDTW$\uparrow$ & CLS$\uparrow$ \\
\hline
1-step mean & 71.94 & 67.21 & 2.69 & 74.28 & 62.05 & 70.98 \\
Adapt-3 mean & 73.25 & 68.72 & 2.59 & 75.66 & 63.77 & 71.98 \\
\hline
$\Delta$ & \textbf{+1.31} & \textbf{+1.51} & \textbf{-0.10} & \textbf{+1.38} & \textbf{+1.72} & \textbf{+1.00} \\
\hline
\end{tabular}
}
\end{table}

\begin{table}[!t]
\renewcommand{\arraystretch}{1.15}
\setlength{\tabcolsep}{4pt}
\caption{Hybrid adaptation with five users on the User (mixed) split. Full-metric comparison against GR-DUET under the same evaluation protocol.}
\label{tab:hybrid_full}
\centering
\resizebox{\columnwidth}{!}{
\begin{tabular}{lcccccc}
\hline
Model & SR$\uparrow$ & SPL$\uparrow$ & NE$\downarrow$ & nDTW$\uparrow$ & SDTW$\uparrow$ & CLS$\uparrow$ \\
\hline
GR-DUET & 61.52 & 56.08 & 3.51 & 66.27 & 51.06 & 64.56\\
Hybrid & 71.43 & 67.22 & 2.72 & 74.48 & 61.96 & 71.18 \\
\hline
\end{tabular}
}
\end{table}

\paragraph{Motivation and setup}
We investigate two complementary user-feedback integration modes that mirror realistic deployment scenarios:
(i) \textbf{Continual adaptation}, where the model is sequentially updated across multiple users over time (only one active user at a time);
and (ii) \textbf{Hybrid adaptation}, where feedback from multiple users is aggregated into a single joint update.

Our goal is not to compare them directly, but to demonstrate that the proposed framework remains effective under both deployment conditions.

\paragraph{Continual adaptation}
In this setting, each experiment involves three sequential updates. 
The first adaptation step uses the \emph{Basic} style (Adapt-1), and the second and third steps (Adapt-2/3) randomly sample from the remaining user styles. 
Between steps, the policy is incrementally updated using newly collected feedback, resulting in approximately 1,500 feedback samples in total. 
Table~\ref{tab:continual_compact} summarizes results across five independent runs. 
Despite varying user instruction distributions, the model maintains stable performance (SR std.\,$\approx$\,0.8, SPL std.\,$\approx$\,0.6), 
demonstrating strong robustness. 
Compared with one-step fine-tuning, continual adaptation consistently improves navigation performance 
(+1.31 SR, +1.51 SPL, and -0.10\,m NE; see Table~\ref{tab:continual_aggregate}), 
indicating that progressive user feedback provides complementary learning signals beyond single-step adaptation. 
Among all sequences, \emph{Basic$\to$Keith$\to$Rachel} achieves the highest SR (74.29) and SPL (69.19), 
while \emph{Basic$\to$Sheldon$\to$Moira} yields the best trajectory alignment with nDTW (76.43) and CLS (72.77).

\begin{table}[t]
\centering
\caption{Training cost and optional memory-bank storage overhead.}
\label{tab:cost_overhead}
\footnotesize
\setlength{\tabcolsep}{6pt}
\renewcommand{\arraystretch}{1.1}
\begin{tabular}{lccc}
\hline
Mode & \#Stages & Train time & \shortstack{Mem (MB)\\(w/o / w/)} \\
\hline
Hybrid (User) & 1 & $\sim$82 h & 0 / 16.32 \\
Continual (Basic+User) & 3 & $\sim$246 h & 0 / $\sim$35 \\
\hline
\end{tabular}
\end{table}

\paragraph{Hybrid adaptation}
In the hybrid setting, feedback from five users (100 samples each, 500 total) is merged into a single adaptation step.
Despite using only one-third of the data and training duration of the continual setup, hybrid adaptation still yields notable improvements over GR-DUET—+9.91 SR, +11.14 SPL, and $-$0.79\,m NE—with consistent gains across nDTW, SDTW, and CLS (Table~\ref{tab:hybrid_full}).
This demonstrates that multi-user feedback aggregation can provide an efficient and scalable solution for real-world deployments where multiple users provide feedback over a deployment window under limited training budgets.

As highlighted by the experimental results, our framework introduces additional components to enable adaptation. To assess the practicality of our method, we analyze the training cost and memory consumption in Table~VI. 
Under identical optimization and hardware settings, each adaptation stage requires approximately 82 hours of training; therefore, Hybrid adaptation performs a single aggregated update within a deployment window and costs 82 hours in total, whereas Continual adaptation updates the agent sequentially across stages and scales roughly linearly with the number of stages (e.g., three stages cost 246 hours). In our reported Continual/Hybrid experiments, we did not enable the persistent memory-bank warm start (save/load across stages), and thus there is no extra persistent storage
overhead (0 MB). The memory-bank warm start is an optional deployment feature, and its storage depends on the set of scans/environments that the agent needs to cache. In our setup, the Basic split covers 24 scans (18.62 MB) while the User split covers 21 scans (16.32 MB). Therefore, Hybrid adaptation (User-only) would
require 16.32 MB if warm start is enabled. For Continual adaptation, we run one stage on Basic scans and two stages on User scans; Since the two User stages share the same scan set but differ in viewpoint coverage, enabling warm start would only require storing the scan set once, totaling approximately 35~MB.

\section{Conclusion}
\label{sec:conclusion}

In this work, we introduced a user-feedback-driven adaptation paradigm for Vision-and-Language Navigation (VLN), fundamentally shifting the reliance from unstable environment-driven self-supervision to intent-aligned human guidance. 
By leveraging a topology-aware trajectory construction pipeline, our framework effectively ``lifts'' sparse, episode-level corrections into dense, grounded supervision, enabling sample-efficient policy refinement without requiring dense step-by-step demonstrations.
We further demonstrated that a persistent, reloadable memory bank significantly mitigates cold-start degradation during redeployment by preserving acquired structural and perceptual knowledge.
Extensive experiments on the GSA-R2R benchmark confirm that our approach consistently outperforms environment-driven baselines on the \textit{standard Basic split}, while effectively handling diverse instruction styles in both \textit{continual and hybrid adaptation settings}. 
This validates that our proposed user feedback mechanism serves as a robust and scalable supervision signal for realistic deployment scenarios.
Although our experiments are conducted in static benchmark environments, the proposed feedback abstraction does not assume scene stationarity. In dynamic settings, the same (success, goal-correction) interface remains valid, while the trajectory lifting step can be performed on an updated topological map.

\bibliographystyle{IEEEtran}
\bibliography{refs}

\end{document}